\def\year{2022}\relax
\documentclass[letterpaper]{article} 
\usepackage{aaai22}  
\usepackage{times}  
\usepackage{helvet}  
\usepackage{courier}  
\usepackage[hyphens]{url}  
\usepackage{graphicx} 
\usepackage{natbib}  
\usepackage{caption} 
\DeclareCaptionStyle{ruled}{labelfont=normalfont,labelsep=colon,strut=off} 
\usepackage{algorithm}
\usepackage{algorithmic}
\usepackage{amsmath,amssymb,amsthm}
\usepackage[english]{babel}
\usepackage{blindtext}
\usepackage{xcolor}

\usepackage{newfloat}
\usepackage{listings}
\floatstyle{ruled}
\newfloat{listing}{tb}{lst}{}
\floatname{listing}{Listing}
\title{Sample-efficient Iterative Lower Bound Optimization of \\ Deep Reactive Policies for Planning in Continuous MDPs}
\author {
    Siow Meng Low\textsuperscript{\rm 1}, 
    Akshat Kumar\textsuperscript{\rm 1}, 
    Scott Sanner\textsuperscript{\rm 2}
}

\affiliations{
    \textsuperscript{\rm 1}School of Computing and Information Systems, 
    Singapore Management University, 
    Singapore \\
    \textsuperscript{\rm 2}Industrial Engineering, 
    University of Toronto, 
    Canada \\

    smlow.2020@phdcs.smu.edu.sg, 
    akshatkumar@smu.edu.sg, 
    ssanner@mie.utoronto.ca
}

\DeclareMathOperator{\tfmdp}{tfmdp}

\DeclareMathOperator{\ilbo}{ILBO}
\usepackage{bibentry}

\newcommand{\bs}{\boldsymbol}

\newcommand{\txn}{\mathcal{T}}

\usepackage{multibib}
\newcites{Appendix}{References for Technical Appendix}

\begin{document}
\maketitle

\begin{abstract}

Recent advances in deep learning have enabled optimization of deep reactive policies (DRPs) for continuous MDP planning by encoding a parametric policy as a deep neural network and exploiting automatic differentiation in an end-to-end model-based gradient descent framework. This approach has proven effective for optimizing DRPs in nonlinear continuous MDPs, but it requires a large number of sampled trajectories to learn effectively and can suffer from high variance in solution quality.  In this work, we revisit the overall model-based DRP objective and instead take a minorization-maximization perspective to iteratively optimize the DRP w.r.t. a locally tight lower-bounded objective.  This novel formulation of DRP learning as iterative lower bound optimization ($\ilbo$) is particularly appealing because (i) each step is structurally easier to optimize than the overall objective, (ii) it guarantees a monotonically improving objective under certain theoretical conditions, and (iii) it reuses samples between iterations thus lowering sample complexity.  Empirical evaluation confirms that $\ilbo$ is significantly more sample-efficient than the state-of-the-art DRP planner and consistently produces better solution quality with lower variance.  We additionally demonstrate that $\ilbo$ generalizes well to new problem instances (i.e., different initial states) without requiring retraining.

\end{abstract}

\section{Introduction}

In the past decade, deep learning (DL) methods have demonstrated remarkable success in a variety of complex applications in computer vision, natural language, and signal processing~\cite{cv17,Hinton12,bengio21}. More recently, a variety of work has sought to leverage DL tools for planning and policy learning in a large variety of deterministic and stochastic decision-making domains~\cite{WuSS17,BuenoBMS19,WuSS20,say20,renato20,Say21,toyer20asnet,garg20generalized}.

A large amount of this work has focused on learning (deep reactive) policies in \emph{discrete} planning settings such as exploiting structure in discrete relational planning models~\cite{fern18drp} or learning policies for effective transfer and generalized planning over different domain instantiations of these relational models~\cite{tamar18drp,toyer18asnet,aniket18transfer,garg19transfer,garg20generalized,toyer20asnet}.
Other recent work has investigated planning by discrete and mixed integer optimization in learned discrete neural network models of planning domains~\cite{SayS18, say20}.
Further afield, policy learning has been 
applied for planning in discrete models through the use of policy gradient methods~\cite{buffet09fpg, buffet07fpg, aberdeen05pgplanning}, although it is critical to remark that these methods focused on model-free reinforcement learning approaches and do not directly exploit optimization over the known model dynamics that we explore in this work.

There has been considerably less research focus in the challenging area of planning in nonlinear \emph{continuous} MDP models that we focus on in this paper.  However, a recent direction of significant influence on the present work is the use of automatic differentiation in an end-to-end model-based gradient descent framework to leverage recent advances in non-convex optimization from DL.  The majority of work in this direction has focused on deterministic continuous planning models --- both known~\cite{WuSS17,renato20} and learned~\cite{WuSS20,Say21}.  However, in this work we are specifically concerned with learning deep reactive policies (DRPs) for fast decision-making in general continuous state-action MDPs (CSA-MDPs).  While there has been work on planning in CSA-MDPs with piecewise linear dynamics via symbolic methods~\cite{ZamaniSF12} or mixed integer linear programming~\cite{sanner17milp}, these methods suffer from scalability limitations and do not learn reusable DRPs for general nonlinear dynamics or reward.  The state-of-the-art solution for DRP learning in such nonlinear CSA-MDPs is provided by $\tfmdp$~\cite{BuenoBMS19}, which optimizes DRPs end-to-end by leveraging gradients backpropagated through the model dynamics and policy.

While $\tfmdp$ made significant advances in solving nonlinear CSA-MDPs, it requires a large number of sampled trajectories to learn effectively and can suffer from high variance in solution quality.  In this work, we revisit the overall model-based DRP objective and instead take a fundamentally different approach to its optimization.  Specifically, we make the following contributions:

\textit{First}, for CSA-MDPs with stochastic transitions, we develop a lower bound for the planning objective for optimizing parameterized DRPs by using techniques from convex optimization and minorize-maximize (MM) methods~\cite{mmBook,Hunter2004}. Exactly optimizing this lower bound is \textit{guaranteed} to increase the original planning objective monotonically. \textit{Second}, we show that this lower bound has a particularly convenient structure for sample-efficient optimization. We also develop techniques that allow us to effectively utilize both 

recent and past data for learning the value function and optimizing the lower bound. This improves sample efficiency and lowers the variance in solution quality. We show that our method can generalize to new problem instances (i.e., different initial states) without retraining. Empirically, we perform evaluation on three different domains introduced in~\cite{BuenoBMS19}. Results confirm that our method \textit{ILBO} is significantly more sample-efficient than the state-of-the-art $\tfmdp$ DRP planner and produces better solution quality with lower variance consistently across all the domains and a variety of hyperparameter settings.

\section{Problem Formulation}
\label{sec:form}

A Markov decision process (MDP) model is defined using the tuple $(S, A, \txn, r, \gamma)$. An agent can be in one of the states $s_t\in S$ at time $t$. It takes an action $a_t\in A$, receives a reward $r(s_t, a_t)$, and the world transitions stochastically to a new state $s_{t+1}$ with probability $p(s_{t+1}| s_t, a_t) \!=\!\txn(s_t, a_t, s_{t+1})$. We assume that rewards are non-negative. For the infinite-horizon setting, future rewards are discounted using a factor $\gamma <1$; for finite-horizon settings $\gamma=1$. The initial state distribution is denoted by $b_0(s)$. We assume continuous, multi-dimensional state and action spaces ($S\subseteq \mathbb{R}^n$, $A\subseteq \mathbb{R}^n$) and denote this as a continuous state-action MDP (CSA-MDP).

\textcolor{black}{The empirical experiments conducted focus on transition function that can be factored},  similar to factored MDPs~\cite{BoutilierDH99}. That is, the transition function can be decomposed as: $p(s_{t+1} | s_t, a_t) \!=\! \prod_{j=1}^n p(s_{t+1}^j|s_t, a_t)$ where $s_{t+1}^j$ is the $j$th component of the state $s_{t+1}$. \textcolor{black}{However, our proposed approach does not require factorization of transition probabilities in general.}

\vskip 2pt
\noindent\textbf{Policy: }We follow a similar setting as in~\cite{BuenoBMS19} and optimize a \textit{deterministic} policy $\mu_\theta$. The Markovian policy $\mu_\theta(s)$ provides the deterministic action $a\!=\!\mu_\theta(s)$ that is to be executed when the agent is in state $s$. The policy $\mu_\theta$ is parameterized using $\theta$ (e.g., $\theta$ may represent parameters of a deep neural net).

\vskip 2pt
\noindent\textbf{Objective: }The state value function is defined as the expected total reward received by following the policy $\mu$ from the state $s_t$ as: $V^\mu(s_t)=\mathbb{E}\Big[\sum_{T=t}^\infty \gamma^{T-t} r(s_T, \mu_\theta(s_T))\Big | s_t) \Big]$. For the finite-horizon setting, we only accumulate rewards until a finite-horizon $H$. The agent's goal is to find the optimal policy $\mu^\star$  maximizing the objective $J$ as:
\begin{align}
 J(\mu) = \mathbb{E}_{s_0\sim b_0}\Big[ V^\mu(s_0) \Big]
\end{align}
where $b_0$ is the initial state distribution.

\section{The Minorize-Maximize (MM) Framework}

Our approach for optimizing DRPs is based on the MM framework; we briefly review it here and refer the reader to~\citeauthor{mmBook}~(\citeyear{mmBook}) for full details. Assume the goal is to solve the optimization problem $\max_{\mu} J(\mu)$. The MM framework provides an iterative approach where at each step $m=0, 1,\ldots$, we construct a function $\hat{J}(\mu; \mu^m)$ (assume that $\mu^0$ is some given starting estimate). The function $\hat{J}(\mu; \mu^m)$ \textit{minorizes} the objective $J(\mu)$ at $\mu^m$ iff:
\begin{align}
\hat{J}(\mu; \mu^m) &\leq J(\mu) \; \forall \mu \label{eq:lb}\\
\hat{J}(\mu^m; \mu^m) &= J(\mu^m) \label{eq:touch}
\end{align}
In the MM framework, we then iteratively optimize the following problem:
\begin{align}
\mu^{m+1} = \arg\max_{\mu} \hat{J}(\mu; \mu^m) \label{eq:max}
\end{align}
The above step is the so-called \textit{maximize} operation in the MM algorithm.
This scheme results in a monotonic increase in the solution quality until convergence (when $\mu^m = \mu^{m+1}$), where we note the minorization property ensures
\begin{align}
J(\mu^{m+1}) \geq \hat{J}(\mu^{m+1}; \mu^m) \, .
\end{align}
Given the maximize operation in~\eqref{eq:max}, we also have $\hat{J}(\mu^{m+1}; \mu^m) \geq \hat{J}(\mu^{m}; \mu^m)$. And we know from condition ~\eqref{eq:touch} that $\hat{J}(\mu^{m}; \mu^m) = J(\mu^m)$. Therefore, $J(\mu^{m+1}) \geq J(\mu^m)$, which shows that the MM algorithm iteratively improves the solution quality until convergence to a local optimum or a saddle point~\cite{mmBook}. 

\begin{figure}[t]
	\centering
	\includegraphics[scale=0.7]{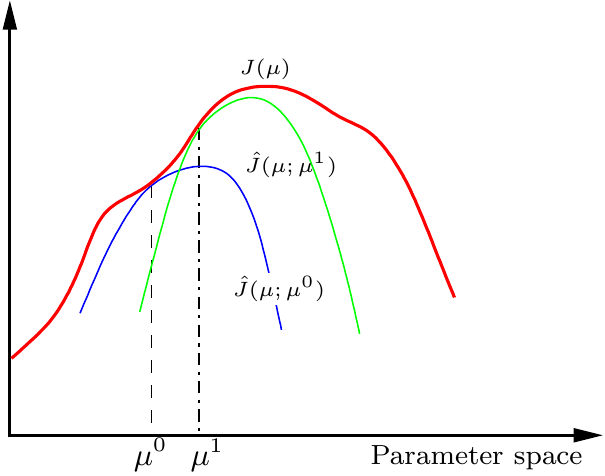}
	\vskip 0pt
	\caption{\small Minorize-Maximize (MM) framework}
	\label{fig:mm}
	\vskip 0pt
\end{figure}

The geometric intuition behind the MM approach is shown in figure~\ref{fig:mm}. We first construct a lower bound $\hat{J}$ of $J$ at the initial estimate $\mu^0$ (or the blue curve $\hat{J}(\mu; \mu^0)$). Then we maximize  $\hat{J}(\mu; \mu^0)$ to get the next estimate $\mu^1$. This step also improves the solution quality as $J(\mu^1) \geq J(\mu^0)$. We then keep repeating this process iteratively by constructing a new lower bound at $\mu^1$ (in green) and maximizing it to get the next estimate until convergence. 

Our goal would be to develop such an MM algorithm for planning in CSA-MDPs.
Several algorithms in machine learning  are based on the MM framework. The well-known expectation-maximization (EM) algorithm~\cite{Dempster1977} can be derived from the  MM perspective, along with other examples noted in~\cite{Hunter2004}. 

In the context of MDPs (and reinforcement learning), several other methods also fall under the umbrella of MM algorithms such as trust region policy optimization (TRPO)~\cite{Schulman2015a}. In contrast to the lower bound in TRPO, our derived lower bounds are much simpler, differentiable, and explicitly involve the gradient of the known reward and transition function in the objective, which is desirable for the planning context. The lower bound in~\cite{Schulman2015a} relies on KL divergence based non-differentiable penalty terms, and requires several approximations to optimize. Furthermore, TRPO assumes a stochastic policy, whereas our lower bound is applicable to the deterministic policy. Lower bounds for the MDP value function have also been proposed when treating planning and RL using a probabilistic inference setting~\cite{Schulman2015,Toussaint2006}. However, such formulations require rewards to not be deterministically influenced by the optimization parameters~\cite{Schulman2015}[Appendix B]. This assumption breaks down for deterministic policies (action $a_t \!=\! \mu(s_t)$), and $r_t = r(s_t, \mu(s_t))$. In contrast, the lower bound we develop is valid for deterministic policies.

\section{MM Formulation For Deterministic Policy}

In Section~\ref{sec:form}, we formulated the objective $J$ for a deterministic policy $\mu$. A lower bound for this objective function using the MM formulation will be derived in this section. The main idea behind \emph{iterative lower bound optimization} ($\ilbo$) is to iteratively improve this lower bound function using a gradient-based method for planning. For ease of exposition, we first consider a discrete state space, and a tabular setting where we need to optimize $\mu(s) \forall s$. Later we generalize this bound for parametric policies and continuous state spaces.

Let $\tau_{0:t}: (s_0, a_0, s_1, a_1, ..., s_t, a_t)$ be a state-action trajectory from $0$ to time $t$. The planning objective $J(\mu)$ can also be expressed as maximizing the following:
\begin{align} 
J(\mu) \!=\! \sum_{T=0}^{\infty} \sum_{\tau_{0:T}}\!\! \gamma^T r(s_T, \mu(s_T)) b_0(s_0)\! \prod_{t=0}^{T-1}\! p(s_{t+1} | s_t, \mu(s_t)) \label{eq:mdpobj}
\end{align}

\subsection{Minorization step in MM}
To fit within the MM framework, our first goal is to minorize the function~\eqref{eq:mdpobj} to create the lower bound $\hat{J}$ satisfying~\eqref{eq:lb} and~\eqref{eq:touch}. At first glance, $J(\mu)$ seems unwieldy due to the presence of terms $\txn(s, \mu(s), s'), r(s, \mu(s))$, which may be an arbitrary nonlinear function of parameters $\mu(s)$. Using tools from convex optimization, we address this challenge. Next, we create additional variables and constraints over them to create a new optimization problem which has a convex objective. 
We first create two sets of additional variables --- $\phi(s, s') \forall s, s'$ and $\omega(s)\forall s$. They encode the transition and reward function in an exponentiated form as:
\begin{align}
e^{\phi(s, s')} = \txn(s, \mu(s), s') \forall s, s', \;\; e^{\omega(s)} = r(s, \mu(s)) \forall s \label{eq:dpg1}
\end{align}
The above representation is always possible as the transition function is non-negative and we assume rewards are also non-negative. Notice that $\mu(s)$ itself is also a variable to optimize. The planning objective $J(\phi, \omega, \mu)$ can be written as: 
\begin{align} 
&\max_{\omega, \phi, \mu} \;  J(\omega, \phi, \mu) = \sum_{T=0}^{\infty} \sum_{\tau_{0:T}} \gamma^T b_0(s_0) e^{\omega(s_T) + \sum_{t=0}^{T-1} \phi(s_t, s_{t+1})} \label{eq:cvx} \\
&e^{\phi(s, s')} = \txn(s, \mu(s), s') \forall s, s', \;\; e^{\omega(s)} = r(s, \mu(s)) \forall s \label{eq:cst}
\end{align}
Notice that the objective in~\eqref{eq:cvx} is convex as it is a positive combination of convex terms (initial belief $b_0$ and $\gamma$ are non-negative). Although policy terms $\mu(s)$ do not appear in the objective, they are related to other variables by constraints~\eqref{eq:cst}. The whole optimization problem is still non-convex as constraints~\eqref{eq:cst} are non-convex (any equality constraints in a convex optimization problem must be linear~\cite{Bertsekas99}). \textcolor{black}{However, this separation is useful as it allows MM procedure to be carried out iteratively. In each iteration, we first perform MM procedure on the objective in~\eqref{eq:cvx} before imposing equality constraints~\eqref{eq:cst} through variable substitution. The later step ensures that the derived solution is confined to the feasible space.} 

Given that $J(\phi, \omega, \mu)$ is convex, we can leverage properties of convex functions to minorize it. Specifically, we use the well-known supporting hyperplane property of a convex function which states that the tangent to the graph of a convex function is a minorizer at the point of tangency~\cite{Hunter2004}. Concretely, if $f(x)$ is convex and differentiable, then:
\begin{align}
f(x) \geq f(x^m) + \nabla f(x^m) \cdot (x - x^m)	\;  \forall x
\end{align}
with equality when $x = x^m$. The minorizer is given as $\hat{f}(x; x^m) = f(x^m) + \nabla f(x^m) \cdot (x - x^m)$. The maximization step in~\eqref{eq:max} is equivalent to:
\begin{align}
\max_{x} 	\nabla f(x^m) \cdot x \label{eq:mmopt}
\end{align}

We have omitted the constant terms that only depend on $x^m$ from the above problem. The analogous  optimization problem to solve in the MM framework for MDPs is given in the next section.

\subsection{Maximization step in MM}

The analogous optimization problem to solve in the MM framework for MDPs is given next. We use the shorthand $J^m = J(\omega^m, \phi^m, \mu^m)$, where superscript $m$ denotes the previous iteration's estimate of the corresponding parameter.  
\begin{align}
	&\hspace{-5pt}\textcolor{black}{\max_{\omega, \phi, \mu}  \sum_{s} \omega(s) \nabla_{\omega(s)} J^m + \sum_{s, s'} \phi(s, s') \nabla_{\phi(s, s')} J^m} \label{obj:dpg}\\
	&\phi(s, s') = \ln \txn(s, \mu(s), s') \; \forall s, s'  \label{cst:dpg1}\\
	&\omega(s) = \ln r(s, \mu(s)) \; \forall s  \label{cst:dpg2}
\end{align}
Notice that the above optimization problem has a much simpler structure than the original problem. The objective function is linear in the parameters to optimize. \textcolor{black}{Note that the minorizer in~\eqref{obj:dpg} does not contain any term dependent on $\mu(s)$ since the objective in~\eqref{eq:cvx} is fully expressed in $\omega(s)$ and $\phi(s, s')$.} The constraints \textcolor{black}{involving $\mu(s)$} are nonlinear, we will show later how to \textcolor{black}{eliminate them through variable substitution.}

The key task to solve problem~\eqref{obj:dpg} is to compute the gradients that are required in the above problem. The analytical expressions for such gradients are derived next. 

\paragraph{Gradient $\nabla_{\omega(s)}J(\omega, \phi, \mu)$}For a given deterministic  policy $\mu$, we can define the occupancy measure for different states as $d^\mu(s)=\sum_{t=0}^\infty \gamma^t p(s_t=s; \mu)$ and $p(\tau_{0:T})$ as probability of trajectory $\tau_{0:T}$. The gradient can be computed as:
\begin{align}
	\frac{\partial J}{\partial \omega(s)} &=  \sum_{T=0}^\infty \sum_{\tau_{0:T}} \gamma^T b_0(s_0) \nabla_{\omega(s)} e^{ \sum_{t=0}^{T-1}  \phi(s_t, s_{t+1})  + \omega(s_T)  } \nonumber \\
	&= \sum_{T=0}^\infty \sum_{\tau_{0:T}} p(\tau_{0:T}) \gamma^T r_T \mathbb{I}_s(s_T)
\end{align}
where $\mathbb{I}_s(s_T)$ is an indicator function returning $1$ iff state at time $T$ is $s$ (or $s_T = s$), otherwise 0. In the above expression, we have also re-substituted $r_T  = e^{\omega(s_T)}$. We further simplify the above to get:
\begin{align}
	&= \sum_{T=0}^\infty \sum_{\tau_{0:T-1}}p(\tau_{0:T-1}) \gamma^T r_T p(s | \tau_{0:T-1}) \\
	&=  \sum_{T=0}^\infty \gamma^T  \sum_{\tau_{0:T-1}} p(\tau_{0:T-1}) p(s | \tau_{0:T-1}) \times r(s, \mu(s))\\
&\frac{\partial J}{\partial \omega(s)}	= d(s) r(s, \mu(s))
\end{align}
where $d(s)$ is the state occupancy measure as defined earlier. Based on probability marginalization we also used:$$\sum_{\tau_{0:T-1}} p(\tau_{0:T-1}) p(s | \tau_{0:T-1}) = p(s_T = s).$$


\paragraph{Gradient $\nabla_{\phi(s, s')}J$}The gradient $\frac{\partial J}{\partial \phi(s, s')}$ is given below:
\begin{align}
\nabla_{\phi(s, s')}J = \gamma d(s) p(s'|s, \mu(s))   V^\mu(s')	
\end{align}
where $d$ is the state occupancy measure. The proof is provided in the supplemental material.

\paragraph{Value function lower bound} Using these gradients, we simplify the problem~\eqref{obj:dpg} as:
\begin{align}
	&\max_{\omega, \phi, \mu}  \sum_{s} d^m(s) r(s, \mu^m(s)) \omega(s) \nonumber \\
	&+ \sum_{s, s'} \gamma d^m(s) \txn(s, \mu^m(s), s')   V^m(s')\phi(s, s') \\
	&\text{subject to constraints~\eqref{cst:dpg1}-\eqref{cst:dpg2}}
\end{align}
where superscript $m$ denotes the dependence of the corresponding term on the previous policy estimate $\mu^m$ (e.g., $V^m$ is the value function of the policy $\mu_{\theta^m}$, and $d^m$ is the corresponding state occupancy measure). We can \textcolor{black}{now }eliminate the equality constraints \eqref{cst:dpg1}-\eqref{cst:dpg2} by substituting them directly into the objective to get the final simplified problem: 
\begin{align}
	&\max_{\omega, \phi, \mu} \hat{J}(\mu; \mu^m) \!=\! \max_{\bs{\mu}} \sum_{s} d^m(s) r(s, \mu^m(s)) \ln r(s, \mu(s)) \nonumber \\
	&+ \!\sum_{s, s'}\!\! \gamma d^m(s) \txn(s, \mu^m(s), s')  V^m(s') \ln \txn(s, \mu(s), s') \label{ccp:dpg}
\end{align}
The function $\hat{J}(\mu; \mu^m)$ is the value function lower bound for the MM strategy shown in figure~\ref{fig:mm}, and each MM iteration is maximizing $\hat{J}$.

There are several ways in which such a MM approach can exploit known model parameters in the planning context. If policy $\mu$ has a simple form (such as a feature based linear policy), then we can directly solve~\eqref{ccp:dpg} using non-linear programming solvers. Notice that we do not have to exactly compute occupancy measures $d^m$; we can replace $\sum_s d^m(s)$ by using expectation over samples collected using previous policy estimate in iteration $m$. Next we focus on optimizing DRPs, which is the parameterized deep neural network based policy used in $\ilbo$.

\paragraph{Optimizing deep reactive policies}For large state spaces, we can parameterize the policy $\mu_\theta$ using parameters $\theta$. In this case, our maximization problem becomes:
\begin{align}
&\max_{\theta} \hat{J}(\theta; \theta^m) \!=\! \max_{\theta} \sum_{s} d^m(s) r(s, \mu_{\theta^m}(s)) \ln r(s, \mu_\theta(s)) \nonumber \\
&+ \!\sum_{s, s'}\!\! \gamma d^m(s) \txn(s, \mu_{\theta^m}(s), s')  V^m(s') \ln \txn(s, \mu_\theta(s), s') \label{eq:vflb}
\end{align}

Since exact maximization over all possible $\theta$ may be intractable, we can optimize $\hat{J}$ by using gradient ascent. Let $\theta^m$ be the current estimate of parameters, the gradient of the lower bound $\nabla_{\theta}\hat{J}|_{\theta=\theta^m}$ is given as (proof in supplement):
\begin{align}
	&\nabla_{\theta}\hat{J}|_{\theta=\theta^m}\!=\! \sum_s \! d^m(s) \! \nabla_\theta \mu_\theta(s) \Big[ \nabla_{\mu(s)}r(s, \mu(s))\big\vert_{\mu(s)= \mu^m(s)} \nonumber\\
	& + \gamma \sum_{s'} \nabla_{\mu(s)} \txn(s, \mu(s), s')\big\vert_{\mu(s)= \mu^m(s)}  V^m(s') \Big]  \label{eq:drp}
\end{align}
Notice that the above expression can be evaluated by collecting samples from current policy estimate $\mu^m$, and computing the gradient of transition and reward functions using autodiff libraries such as Tensorflow. Iteratively optimizing the above expression results in our iterative lower bound optimization ($\ilbo$) algorithm. In addition, we also develop a principled method to reduce the variance of gradient estimates (shown in supplement) without introducing any bias. Note  that the state value function $V^m(s)$ is not known and has to be estimated. $\ilbo$ utilizes a deep neural net to learn state-action approximator ${\hat{Q}}_\psi(s, a)$ and estimates it using the relation ${\hat{V}}^{m}(s) = {\hat{Q}}_\psi(s, a) |_{a = {\mu^m}(s)}$.

\begin{figure*}[htb]
	\centering
	\includegraphics[width=1.0\textwidth]{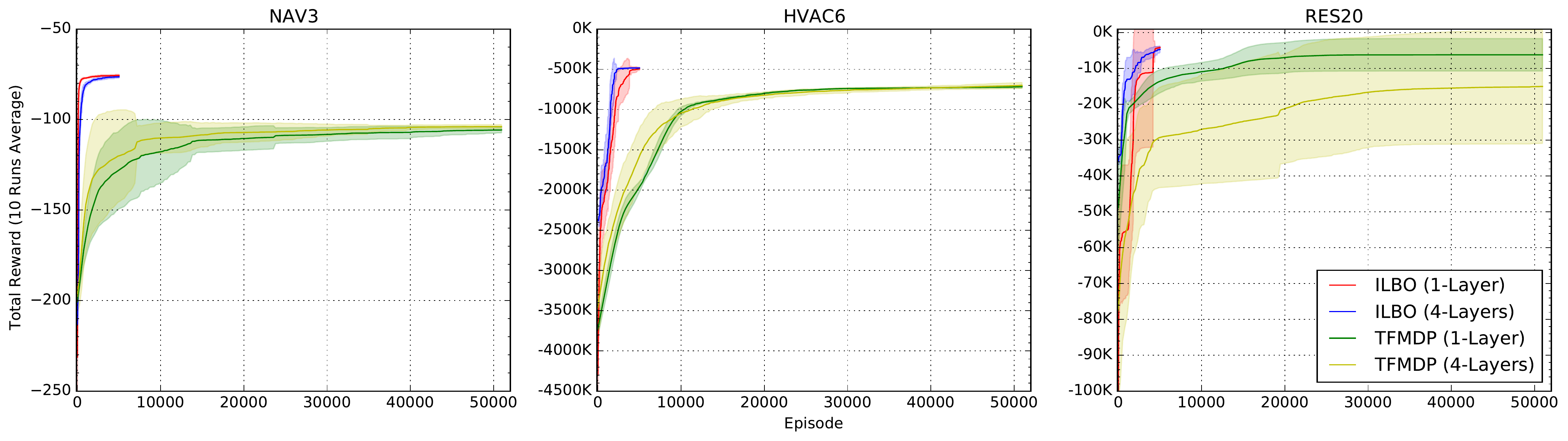}
	\caption{\small Total Reward Obtained (Higher is Better); $\tfmdp$ setting: 200 epochs; 256 Batchsize. x-axis denotes \# episodes, y-axis quality}
	\label{fig:Combined_train_best}
\end{figure*}

We also remark on the structure of expression~\eqref{eq:drp}. Gradient ascent tries to optimize the reward in the first term. In the second term, it tries to increase the probability of transitioning to highly-valued states. Since approximator $\hat{V}$ informs $\ilbo$ of the highly-valued states, gradient ascent on the transition function adjusts the policy such that it makes transition to highly rewarding states more likely. And the use of function approximator for state-action value function smooths out the effect of environment stochasticity as action-value function can be learned from past experiences also (explained later). These features of $\ilbo$ contribute to the sample efficiency and better quality solutions than $\tfmdp$. Another key difference between $\ilbo$ and $\tfmdp$ is that $\ilbo$ uses gradient of the transition function directly, whereas $\tfmdp$ encodes this information implicitly with reparameterization trick~\cite{BuenoBMS19}.

\vskip 2pt
\noindent{\textbf{Extension to continuous state spaces: }}Previous sections derived the MM formulation for a discrete state space MDP. We can also show how the MM formulation extends easily to the continuous state space also using Riemann sum based approximation of integrals in the continuous state space setting (proof in the supplemental material). The structure of the lower bound even in the continuous state setting remains the same as in~\eqref{eq:vflb} with summation over states replaced using an expectation over sampled states from the policy $\mu^m$.

\vskip 2pt
\noindent{\textbf{Efficient utilization of samples: }}Notice that computing the gradient~\eqref{eq:drp} requires on-policy samples, similar to $\tfmdp$. Therefore, we maintain a small store of recently observed episodes and use it to optimize the lower bound. However, our approach can still utilize samples from the past (or the so-called \textit{off-policy}) samples to effectively train the state value estimate ${\hat{V}}_\psi(s)$ of the current policy. This makes our approach much more stable and sample efficient than $\tfmdp$ as our value function estimate is intuitively more accurate than  estimating the value of a state only using on-policy samples from the current mini-batch of episodes as in $\tfmdp$. We provide more on such implementation details, and the entire $\ilbo$ algorithm in the supplementary.

\section{Experiments}

We present the empirical results comparing \textcolor{black}{performance of $\ilbo$\footnote{\url{https://github.com/siowmeng/ilbo}}} against \textcolor{black}{that of} state-of-the-art DRP planner, $\tfmdp$~\cite{BuenoBMS19}. Experiments were carried out in the three planning domains used to evaluate $\tfmdp$ \cite{BuenoBMS19}. We only provide a brief description of these three domains here; for details we refer to~\cite{BuenoBMS19}. Experiments in these three planning domains have been conducted using the Python simulator made publicly available on the RDDLGym GitHub repository~\cite{rddlgym2020}. For $\tfmdp$, we used its Python implementation, available on GitHub~\cite{tfmdp2021}.

\textbf{Navigation} is a path planning problem in a two-dimensional space \cite{faulwasser2009nonlinear}. The agent's location is encoded as a continuous state variable \(\boldsymbol{s}_t \in \mathbb{R}^2\) while the continuous action variable \(\boldsymbol{a}_t \in [-1, 1]^2\) represents its movement magnitudes in the horizontal and vertical directions. The objective is to reach the goal state in the presence of deceleration zones.

\textbf{HVAC} control is a centralized planning problem where an agent controls the heated air to be supplied to each of the $N$ rooms~\cite{10.1145/1878431.1878433}. Continuous state variable \(\boldsymbol{s}_t \in \mathbb{R}^N\) represents the temperature of each room and action variable \(\boldsymbol{a}_t \in [0, a_{max}]^N\) is comprised of the amount of heated air supplied; $N=6$ in this domain. The objective is to maintain the room temperatures within the desired range.

\textbf{Reservoir} Reservoir management requires the planner to release water to downstream reservoirs to maintain the desired water level~\cite{Yeh1985ReservoirMA}. State variable ${s}_t^i$ \,is the water level at reservoir $i$ and action variable ${a}_t^i \in [0, {s}_t^i]$ represents the water outflow to the downstream reservoir. A penalty is imposed if the reservoir level is too low or high. Both state and action spaces are 20-dimensional.

\begin{figure*}[ht]
	\centering
	\includegraphics[width=1.0\textwidth]{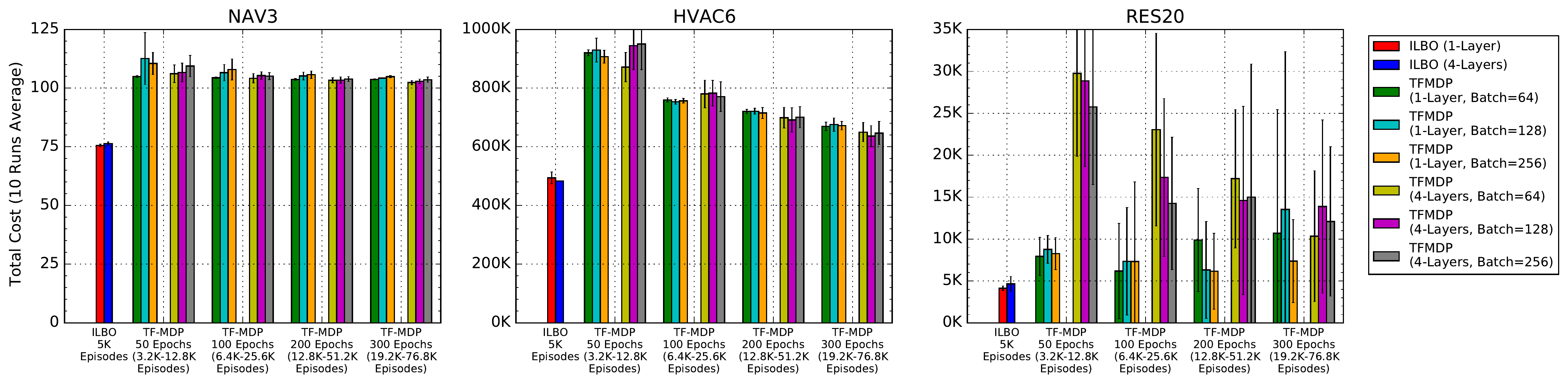}
	\caption{\small Total Cost Incurred by the Final Policy (Lower is Better); $\tfmdp$: Various Hyperparameter Settings}
	\label{fig:Combined_bars}
\end{figure*}

\textbf{Experiment Objectives} We assess $\ilbo$'s performance in the following aspects:

\begin{itemize}
\item Sample efficiency and quality of solution: The ability to provide high-quality solution (in terms of total rewards gathered) in a sample-efficient manner.
\item Training stability: The policy network should gradually stabilize and not exhibit sudden large degradations in performance as learning progresses.
\item Generalization capability: The ability to generalize to other problem instances (e.g., new initial states) without having to retrain.
\end{itemize}

\begin{figure}[t]
	\centering
	\includegraphics[scale=0.4]{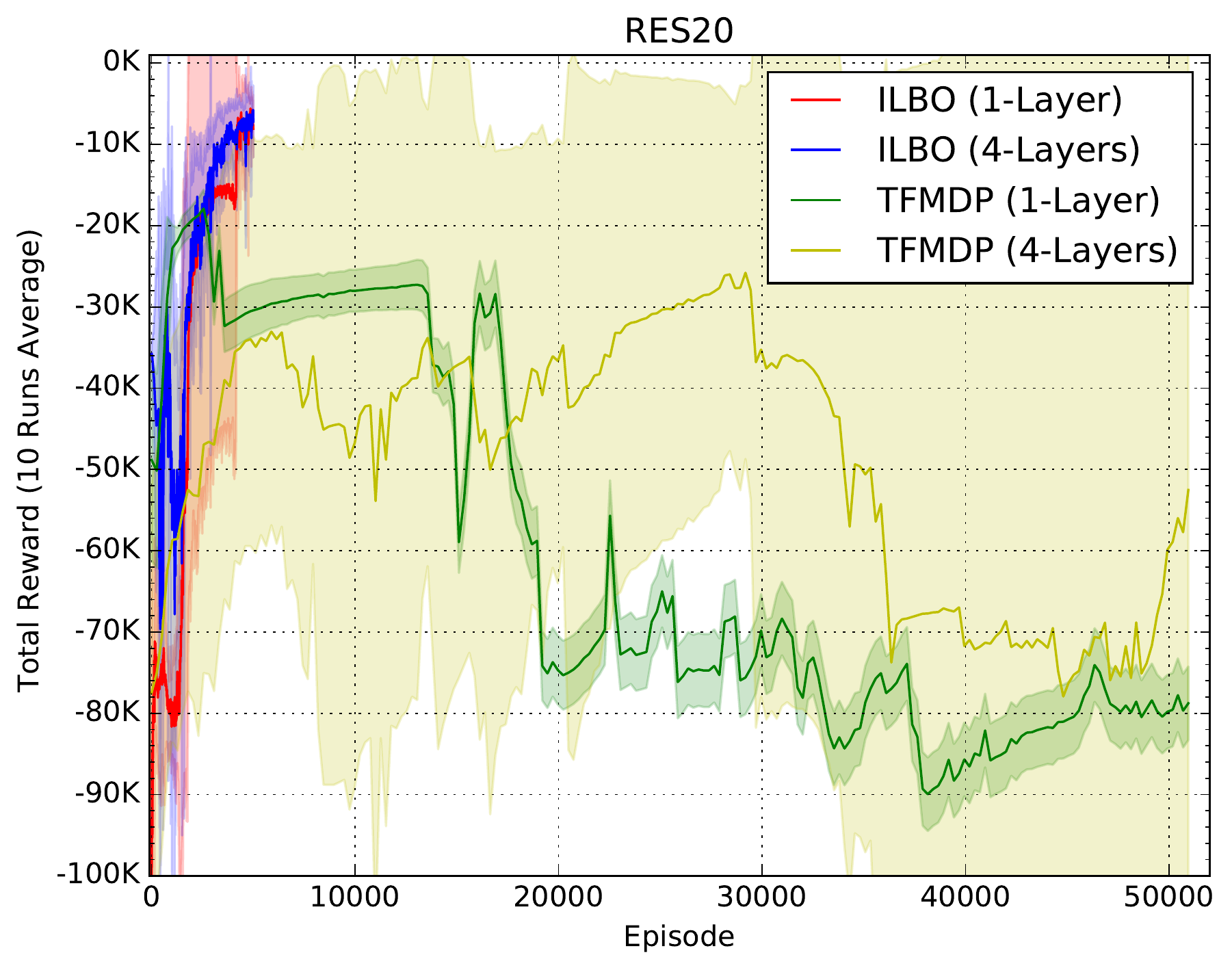}
	\vskip 0pt
	\caption{\small Total Reward Obtained by Current Policy in RES20 (Higher is Better); $\tfmdp$ setting: 200 epochs \& 256 Batchsize}
	\label{fig:res20_train_avg}
	\vskip 0pt
\end{figure}

\textbf{Hyperparameters} We evaluated $\ilbo$ using two representative DRP architectures: one hidden layer and four hidden layers, exactly the same architectures used in the evaluation~\cite{BuenoBMS19}. The total learning episodes for $\ilbo$ is 5000 in all our experiments. The default setting for $\tfmdp$ is: 200 epochs of 256 batchsize, where batchsize refers to the number of episodes (or trajectories) sampled per epoch~\cite{BuenoBMS19}. We performed detailed comparisons with a variety of $\tfmdp$ hyperparameter settings: epoch = \{50, 100, 200, 300\}, batchsize = \{64, 128, 256\}. We refer the reader to supplement for other hyperparameter settings used in the experiments.

\textbf{Performance Reporting} We adopt the same approach as $\tfmdp$ ~\cite{BuenoBMS19}, which reports the average and standard deviation values over 10 training runs and for each training run, the best policy so far is kept to smooth out the effects of stochasticity during training. All the result figures report solution qualities attained by the best policies so far, with the exception of Figure~\ref{fig:res20_train_avg}, which compares the training stability of $\ilbo$ and $\tfmdp$. We use 64 test trajectories (different from training ones) to evaluate the  policy. 

\textbf{Sample Efficiency, Quality} The total rewards gathered by $\ilbo$ and $\tfmdp$ are shown in Figure~\ref{fig:Combined_train_best} for all three domains. The two $\tfmdp$ curves were trained using the default settings specified in \cite{BuenoBMS19}: epoch = 200, batchsize = 256. Compared to $\tfmdp$, $\ilbo$ converges to higher quality plan earlier and collects much fewer sample episodes in all three domains. There are several key reasons for this. Our method can learn better estimate of state value function using off-policy samples, which intuitively is more accurate than estimating value function from only on-policy samples. For highly stochastic domains, such as reservoir, limited on-policy samples may not be enough to provide a reliable estimate of the value function. Additionally, the gradient in~\eqref{eq:drp} can be estimated after collecting sufficient experience samples, without having to wait for full trajectory samples, which is required in $\tfmdp$. Consequently, $\ilbo$ is able to produce high quality plans in a sample-efficient manner over $\tfmdp$, as the empirical result suggests.

\textbf{Training Stability} From Figure~\ref{fig:Combined_train_best}, the reader may notice the large variance present in the total rewards gathered by $\tfmdp$ for Reservoir-20 domain. Compared to the other two domains, Reservoir-20 domain is highly stochastic and its transition dynamics is characterized by the Gamma distribution, which has fat tails at both ends. Consequently, each mini-batch of collected samples can be highly dissimilar. The effect of this high stochasticity on $\ilbo$ and $\tfmdp$ is further examined in Figure \ref{fig:res20_train_avg}. Instead of remembering the best policy so far, this diagram shows the average total reward of current policies at every training step. 

Figure \ref{fig:res20_train_avg} demonstrates that $\ilbo$ is more resilient to environment stochasticity. The weight updates by $\ilbo$ are more stable and the policy improves iteratively. In comparison, $\tfmdp$ witnesses large variations in policy performance towards the later part of training. We postulate that this is due to the better learning of value estimates by $\ilbo$. Although the environment is highly stochastic, $\ilbo$ updates value function using historical samples. As a result, this stochastic effect was smoothed out. In contrast, $\tfmdp$ only uses the current batch of trajectory samples to compute gradient updates. \textcolor{black}{In a highly stochastic domain like Reservoir-20, one single batch might not contain sufficient number of representative trajectory samples. Consequently, $\tfmdp$ might estimate gradients using these unrepresentative samples, causing huge performance degradation. } 

\textbf{Varying hyperparameters of $\boldsymbol{\tfmdp}$} The bar charts in Figure \ref{fig:Combined_bars} quantify the total cost achieved at the end of training. This chart presents the full range of $\tfmdp$ performance across a variety of epoch and batchsize settings, in order to provide an elaborate comparison. Note that cost is defined as negative reward and hence lower is better in this figure. 

\begin{figure*}[t]
	\centering
	\includegraphics[width=1.0\textwidth]{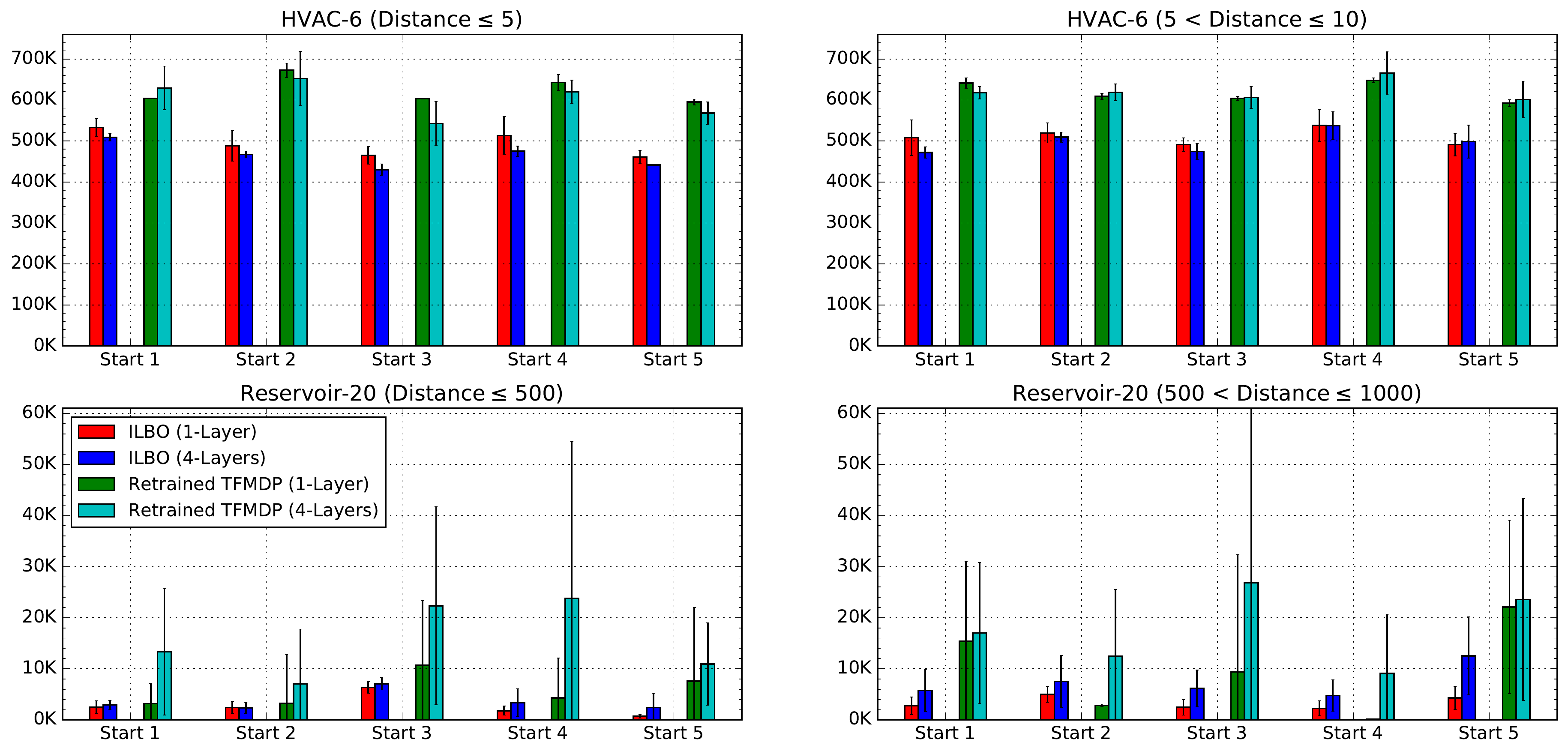}
	\caption{\small Total Cost Incurred in New Initial States (Lower is Better): Same $\ilbo$ Policy vs Retrained $\tfmdp$ Policy}
	\label{fig:Combined_new_init_states}
\end{figure*}

In HVAC and Reservoir domains, increasing the number of epochs and/or batchsize improves the solution qualities attained by $\tfmdp$, albeit at the cost of sample efficiency. We continue to witness high variance in $\tfmdp$'s Reservoir-20 policy performance; this corroborates our finding in Figure~\ref{fig:res20_train_avg}. The best $\tfmdp$ policy in Reservoir domain is trained with 200 epochs of 256 batchsize. However, its high variance indicates that $\tfmdp$ often produces significantly worse policies in many of the runs.

In comparison, $\ilbo$ consistently produces higher quality solutions across all three domains, with significantly lower average total cost and variance. It is worth pointing our that even in the highly stochastic Reservoir domain, the performance variance of $\ilbo$ policies remain low. This is advantageous since it provides the assurance that $\ilbo$ consistently produces policies with similar performance, despite high stochasticity present in the environment. 

\textbf{Generalization} The $\tfmdp$ algorithm estimates gradients from a fixed-length Stochastic Computation Graph \cite{Schulman2015} and the planning horizon is fixed before training commences. Coupled with fixed initial state, the training samples collected by agent may be too monotonous and disadvantage the DRP in generalizing to new states. In contrast, the better training of state value approximator coupled with explicitly taking gradients of reward and transition function in $\ilbo$~\eqref{eq:drp} helps $\ilbo$ to ride out the idiosyncrasy present within a single minibatch of samples, and generalize better to new states.

Figure \ref{fig:Combined_new_init_states} demonstrates $\ilbo$'s ability to generalize. We evaluated the same $\ilbo$ policies trained for the original start state in HVAC and Reservoir domains, without performing retraining. Ten different random start states were used in this evaluation. Five of them are within shorter Euclidean distance from the original start state while the other five are of larger distance. We retrained $\tfmdp$ in these new start states to serve as benchmark for comparison.

For HVAC-6 domain, $\ilbo$ consistently outperforms the retrained $\tfmdp$ in all ten start states. As for Res-20 domain, $\ilbo$ outperforms the retrained $\tfmdp$ in all but two initial states (i.e., Start 2 and Start 4 in the lower right chart). We postulate that this is because the state space is high-dimensional and very large in Res-20 domain, and the agent may not have collected sufficient representative transition samples within 5000 episodes. More training episodes would allow $\ilbo$ to improve generalization in the learned policy. 

\section{Conclusion and Future Work}

In this work, we developed a novel minorization-maximization approach for sample-efficient deep reactive policy (DRP) learning in nonlinear CSA-MDPs using \emph{iterative lower bound optimization} ($\ilbo$).  Empirical results confirmed $\ilbo$'s superior sample efficiency, solution quality, stability, and generalization to new initial states by learning from diverse transition samples. 

One interesting direction of future work would be to extend $\ilbo$ to other types of MDP problems such as constrained MDPs \cite{altman1999constrained, feyzabadi2014risk}.  Solving constrained MDPs in a sample-efficient manner is particularly important in many safety-critical settings germane to the deployment of DRPs. Another promising direction would be to develop $\ilbo$ extensions for optimizing DRPs that are robust to model error \cite{NE:05, wiesemann2013robust}, which would further facilitate their practical deployment when high fidelity models are hard to learn or acquire. 

\section*{Acknowledgments}

We thank anonymous reviewers for their helpful feedback. This research/project is supported by the National Research Foundation, Singapore under its AI Singapore Programme (AISG Award No: AISG2-RP-2020-017).

{
\small 
\bibliography{aaai22MM}}

\clearpage

\end{document}


\onecolumn
\begin{center}
    \LARGE\textbf{Supplemental Materials}
\end{center}

\section{Technical Appendix}

\setcounter{equation}{24}
\setcounter{figure}{5}

\subsection{Proof - Gradient of MDP Objective}

\paragraph{Gradient $\nabla_{\phi(s, s')}J$} The gradient $\frac{\partial J}{\partial \phi(s, s')}$ is derived as:
\begin{align}
	\frac{\partial J}{\partial \phi(s, s')} &=  \sum_{T=0}^\infty \sum_{\tau_{0:T}} \gamma^T b_0(s_0) \nabla_{\phi(s, s')} e^{ \sum_{t=0}^{T-1}  \phi(s_t, s_{t+1})  + \omega(s_T)  } \nonumber \\
	&= \sum_{T=0}^\infty \sum_{\tau_{0:T}} p(\tau_{0:T}) \gamma^T r_T \sum_{t=0}^{T-1} \mathbb{I}_{s, s'}(s_t, s_{t+1}) \nonumber \\
	&= \sum_{T=0}^\infty \sum_{t=0}^{T-1} \sum_{\tau_{0:T}} p(\tau_{0:T}) \gamma^T r_T \mathbb{I}_{s, s'}(s_t, s_{t+1}) \nonumber \\
	&= \sum_{T=0}^\infty \sum_{t=0}^{T-1} \sum_{\tau_{0:t-1}} \gamma^t p(\tau_{0:t-1}) p(s | \tau_{0:t-1}) p(s'|s, \mu(s)) \sum_{\tau_{t+2:T}} \gamma^{T-t} p(s', \mu(s'), \tau_{t+2:T}) r_T
\end{align}
In the third step of above derivation, $\sum_{\tau_{0:T}} p(\tau_{0:T}) \gamma^T r_T \mathbb{I}_{s, s'}(s_t, s_{t+1})$ refers to the sum of all trajectories passing through $s_t = s$ and $s_{t+1} = s'$. In the fourth step, we simply break the trajectories into two collections of trajectories: $\tau_{0:t-1}$ and $\tau_{t+2:T}$ with $s_t = s$, $s_{t+1} = s'$. 

The term $p(s', \mu(s'), \tau_{t+2:T})$ is the probability of a trajectory starting at $s_{t+1} = s'$ following policy $\mu$. Note that the sum $\sum_{\tau_{t+2:T}} \gamma^{T-t} p(s', \mu(s'), \tau_{t+2:T}) r_T \;$ averages the discounted reward across all trajectories from timestep $t+1$ to timestep $T$. We can express this aggregation as the expected discounted reward after $T-t-1$ timesteps: $\mathbf{E}_{s_0 = s', a \sim \mu} [\gamma^{T-t} r_{T-t-1}]$.

\begin{align}
	\frac{\partial J}{\partial \phi(s, s')} &= \sum_{T=0}^\infty \sum_{t=0}^{T-1} \sum_{\tau_{0:t-1}} \gamma^t p(\tau_{0:t-1}) p(s | \tau_{0:t-1}) p(s'|s, \mu(s)) \mathbf{E}_{s_0 = s', a \sim \mu} \Big[\gamma^{T-t} r_{T-t-1}\Big] \nonumber \\
	&= \sum_{T=0}^\infty \sum_{t=0}^{T-1} \mathbf{E}_{s_0 = s', a \sim \mu} \Big[\gamma^{T-t} r_{T-t-1}\Big] \gamma^t p(s'|s, \mu(s)) \sum_{\tau_{0:t-1}} p(\tau_{0:t-1}) p(s | \tau_{0:t-1})   \nonumber \\
	&= \sum_{t=0}^{\infty} \sum_{T=t+1}^{\infty} \mathbf{E}_{s_0 = s', a \sim \mu} \Big[\gamma^{T-t} r_{T-t-1}\Big] \gamma^t p(s'|s, \mu(s)) p(s_t = s; \mu) \nonumber \\
	&= \sum_{t=0}^{\infty} \gamma^t p(s_t = s; \mu) p(s'|s, \mu(s)) \sum_{T=t+1}^{\infty} \mathbf{E}_{s_0 = s', a \sim \mu} \Big[\gamma^{T-t} r_{T-t-1}\Big] \nonumber \\
	&= \sum_{t=0}^{\infty} \gamma^{t+1} p(s_t = s; \mu) p(s'|s, \mu(s)) \sum_{\alpha=0}^{\infty} \mathbf{E}_{s_0 = s', a \sim \mu} \Big[\gamma^{\alpha} r_{\alpha}\Big] \nonumber \\
	&= \gamma p(s'|s, \mu(s)) V^{\mu}(s') \sum_{t=0}^{\infty} \gamma^{t} p(s_t = s; \mu) \nonumber \\
	&= \gamma d(s) p(s'|s, \mu(s))   V^\mu(s')
\end{align}

In the above derivation, we used the same probability marginalization rule $\sum_{\tau_{0:T-1}} p(\tau_{0:T-1}) p(s|\tau_{0:T-1}) = p(s_T=s)$ introduced in the main text. We also performed a change of variable $\alpha = T-t-1$ and derived the last step using the definition of state occupancy measure $d(s)$.

\subsection{Proof - Gradient of Value Function Lower Bound}

The gradient of $\hat{J}$ in~(\vflb) wrt $\theta$ is derived as:
\begin{align} \label{eq:grad_vflb}
 \frac{\partial \hat{J}}{\partial \theta} &= \sum_{s} d^m(s) r(s, \mu_{\theta^m}(s)) \nabla_{\theta} \ln r(s, \mu_\theta(s)) + \sum_{s, s'} \gamma d^m(s) \txn(s, \mu_{\theta^m}(s), s')  V^m(s') \nabla_{\theta} \ln \txn(s, \mu_\theta(s), s') \nonumber \\
 &= \sum_s \! d^m(s) \! \nabla_\theta \mu_\theta(s) \Big[ r(s, \mu_{\theta^m}(s)) \frac{\nabla_{\mu(s)} r(s, \mu(s))}{r(s, \mu(s))} + \gamma \sum_{s'} \txn(s, \mu_{\theta^m}(s), s') \frac{\nabla_{\mu(s)} \txn(s, \mu(s), s')}{\txn(s, \mu(s), s')} V^m(s') \Big]
\end{align}

Evaluating~(\eqGrad) 
at $\theta = {\theta}^m$ gives the expression in~(\drp).

\subsection{Proof - Reduce Gradient Variance Using Baseline Subtraction}

We developed a principled method which is observed to greatly reduce variance of the estimated gradient in our empirical experiments. This method adds subtraction term to~(\drp):

\begin{align} \label{eq:grad_baseline}
	& \nabla_{\theta}\hat{J}|_{\theta=\theta^m} = \sum_s \! d^m(s) \! \nabla_\theta \mu_\theta(s) \Big[ \nabla_{\mu(s)}r(s, \mu(s))\big\vert_{\mu(s)= \mu^m(s)} + \gamma \sum_{s'} \nabla_{\mu(s)} \txn(s, \mu(s), s')\big\vert_{\mu(s)= \mu^m(s)}  (V^m(s') - V^m(s)) \Big]
\end{align}

The added subtraction term does not alter gradient since it is equal to zero when evaluated at $\theta = {\theta}^m$:

\begin{align} \label{eq:grad_baseline_zero}
 & -\sum_s \! d^m(s) \! \nabla_\theta \mu_\theta(s) \gamma \sum_{s'} \nabla_{\mu(s)} \txn(s, \mu(s), s')\big\vert_{\mu(s)= \mu^m(s)} V^m(s) \nonumber \\
 &= - \gamma \sum_s \! d^m(s) \! \nabla_\theta \mu_\theta(s) V^m(s) \nabla_{\mu(s)} \Big[ \sum_{s'} \txn(s, \mu(s), s')\big\vert_{\mu(s)= \mu^m(s)} \Big] \nonumber \\
 &= - \gamma \sum_s \! d^m(s) \! \nabla_\theta \mu_\theta(s) V^m(s) \nabla_{\mu(s)} \big( 1 \big) \nonumber \\
 &= 0
\end{align}

\subsection{Proof - Extension of MM Formulation to Continuous State Space}

In the main text, we derived the MM formulation for a discrete state space MDP. We now show how MM formulation extends easily to the continuous state spaces also. We use Riemann sum based approximation of integrals, which is suitable for MM formulation for CSA-MDPs. We assume a deterministic policy $\mu$. For exposition clarity, we assume single-dimensional continuous states $s\in[a, b]$; derivation below can be extended to multidimensional continuous state spaces also.
The objective is:
{
	\begin{align}
		J(\mu) = \sum_{T=0}^\infty \int_a^b b_0(s_0) \Big[\int_a^b \txn(s_0, \mu(s_0), s_1) \Big(\ldots \int_a^b \txn(s_{T-1}, \mu(s_{T-1}), s_T) r(s_T, \mu(s_T)) d_{s_T} \Big) d_{s_1} \Big] ds_0
\end{align}}
To employ Riemann sum approximation, we divide the interval $[a, b]$ into $n$ partitions as $\mc{P} = \{[x_0, x_1], \ldots, [x_{n-1}, x_n]   \}$. Let us now consider approximating the outermost integral $\int_{a}^b  b_0(s_0) f(s_0) d_{s_0}$ where $f_0(s_0)$ is the expression:
{
	\begin{align*}
		f_0(s_0) \!=\!\! \int_a^b \!\! \txn(s_0, \mu(s_0), s_1) \Big(\ldots \int_a^b \txn(s_{T-1}, \mu(s_{T-1}), s_T) 
		r(s_T, \mu(s_T)) d_{s_T} \Big) d_{s_1}
\end{align*}}
Using Riemann sum, it is approximated as:
\begin{align}
	\int_{a}^b  b_0(s_0) f_0(s_0) d{s_0} \approx \sum_{i=1}^n b_0(x_{i}^\star) f_0(x_{i}^\star) \Delta    
\end{align}
Where $x_{i}^\star$ is some point in the interval $[x_{i-1}, x_i]$; and $\Delta$ is $(x_{i+1} - x_i)$ (for simplified notation, we assume equal partitions). Using this reasoning recursively, we can approximate $J(\mu)$ as $\tilde{J}(\mu)$:
{
	\begin{align*}
		\tilde{J}(\mu) = \sum_{T=0}^\infty \sum_{i=1}^n b_0(x_i^\star)\Delta \sum_{i_1=1}^n \txn(x_{i}^\star,\mu(x_{i}^\star), x_{i_1}^\star ) \Delta \ldots 
		\sum_{i_T=1}^n \txn(x_{i_{T-1}}^\star, \mu(x_{i_{T-1}}^\star),x_{i_T}^\star ) \Delta\; r(x_{i_T}^\star, \mu(x_{i_T}^\star)) \gamma^T
\end{align*}}

\begin{algorithm}[thb]
\caption{ILBO algorithm}
\label{algS:algorithm}
\textbf{Models}: $Q(s, a | \psi)$ and $\mu(s | \theta)$
\begin{algorithmic}[1] 
\small
\STATE Randomly initialize weights $\psi$\, and\, $\theta$
\STATE Copy weights $\psi$\, and\, $\theta$ to target network weights $\psi'$\ and $\theta'$
\STATE Initialize sample stores $E^Q$\, and\, $E^{\mu}$
\FOR{episode $= 1, 2, ... K$}
\STATE Sample start state $s_0$ from feasible state space
\WHILE{not terminal state}
\STATE $\mathcal{N}_t \gets $ Diagonal Gaussian Noise with decay
\STATE Perform $a_t = \mu(s_t | \theta) + \mathcal{N}_t$ and observe transition
\STATE Store experience $(s_t, a_t, r_t, s_{t+1}$) in $E^Q$\, and\, $E^{\mu}$
\STATE Sample a random minibatch (size $n$) from $E^Q$
\STATE Set $y_i = r_i + \gamma Q'(s_{i+1}, {\mu}'(s_{i+1} | \theta') | \psi')$
\STATE Update $\psi$ by minimizing: $\frac{1}{n} \sum_{i} {(y_i - Q(s_i, a_i | \psi)}^2$
\STATE Sample a random minibatch (size $m$) from $S^{\mu}$
\STATE Update $\theta$ using gradient estimated from~(\eqBase)
\STATE Update target networks:
\begin{align*}
 \psi' \gets \tau \psi + (1 - \tau) \psi' \\
 \theta' \gets \tau \theta + (1 - \tau) \theta'
\end{align*}
\ENDWHILE
\ENDFOR
\STATE \textbf{return} $Q(s, a | \psi)$ and $\mu(s | \theta)$
\end{algorithmic}
\end{algorithm}

Notice that the above expression is identical to the value function of a discrete state space MDP with states $S\!=\! \{x_i^\star \forall i\!=\!1
\!:\!n\}$, transition function $\tilde{T}(x_i^\star, a, x_j^\star) \!=\! \txn(x_i^\star, a ,x_j^\star ) \Delta$, and initial state distribution $\tilde{b}_0(x_i^\star) \!=\! b_0(x_i^\star)\Delta$. Once we have such a discretization, we can apply techniques developed in the main text for discrete state space MDP. The lower bound expression and the gradient-based updates of lower bound would be similar as for the discrete state MDP. A key point to note is that the summation $\sum_s d^m(s)$ is replaced with expectation over state samples generated from state occupancy distribution of the previous iteration's policy. We note that the implementation \textit{never} actually discretizes the state space, the discretization viewpoint helps in the derivation; final updates are implementable by generating samples from occupancy distribution $d^m(s)$.

\subsection{Implementation Details of $\ilbo$ Algorithm}

This section outlines the components used in $\ilbo$ to improve deterministic policy $\mu_{\theta}$. The detailed steps of the entire $\ilbo$ learning algorithm are described in Algorithm~\alg.

\textbf{State Action Approximator} The state-action approximator $\hat{Q}_{\psi}(s, a)$ the adopts the mean-square error loss function specified in \citeSupplement{ap_DBLP:journals/corr/MnihKSGAWR13}. To use the mean-square error loss, two target networks $Q'$ and ${\mu}'$ are required. $\ilbo$ performs soft target updates \citeSupplement{ap_Lillicrap2016} to update the weights of these two target networks slowly, with a small value of update rate $\tau$ in the target network update step of Algorithm~\alg.

\textbf{Sample Store} For $\ilbo$, both state action approximator and DRP learn from experience samples. To facilitate learning, two sample stores $E^Q$ and $E^{\mu}$ are used to record experience samples. The sample store concepts were inspired by~\citeSupplement{ap_DBLP:journals/corr/MnihKSGAWR13}, which introduces random sampling of experiences in order to break correlation between samples. These two sample stores maintained by $\ilbo$ are of different sizes.

The sample store for state action approximator $E^Q$ is larger and it holds a larger set of experiences. Since state action approximator can perform learning using off-policy samples, its sample store can be larger to encourage reuse of past samples for better generalization. For DRP, its sample store $E^{\mu}$ is of much smaller size. This is because its gradient has to be estimated using on-policy experience samples, as specified in Equation~(\eqBase). 
For this purpose, $\ilbo$ maintains a much smaller sample store $E^{\mu}$ for the DRP so that the samples drawn are approximately on-policy. This approximation holds when the policy network is updated slowly; hence the recent experiences are approximately distributed according to current policy. 

Learning the state action approximator $\hat{Q}_{\psi}(s, a)$ using past samples drawn from the sample store helps improve the sample efficiency of the learning process. Intuitively, the state action approximator needs to collect fewer experience samples to approximate $Q(s, a)$ due to its ability to learn from all past off-policy samples. In addition, it reduces overfitting and makes the approximations more generalizable across all seen experiences. As a result, the DRP $\mu_{\theta}(s)$ benefits from better estimate of ${\hat{V}}^{m}(s)$ and can better adjust the policy towards transitioning to high value states.

\begin{figure}[ht]
	\centering
	\includegraphics[scale=0.5]{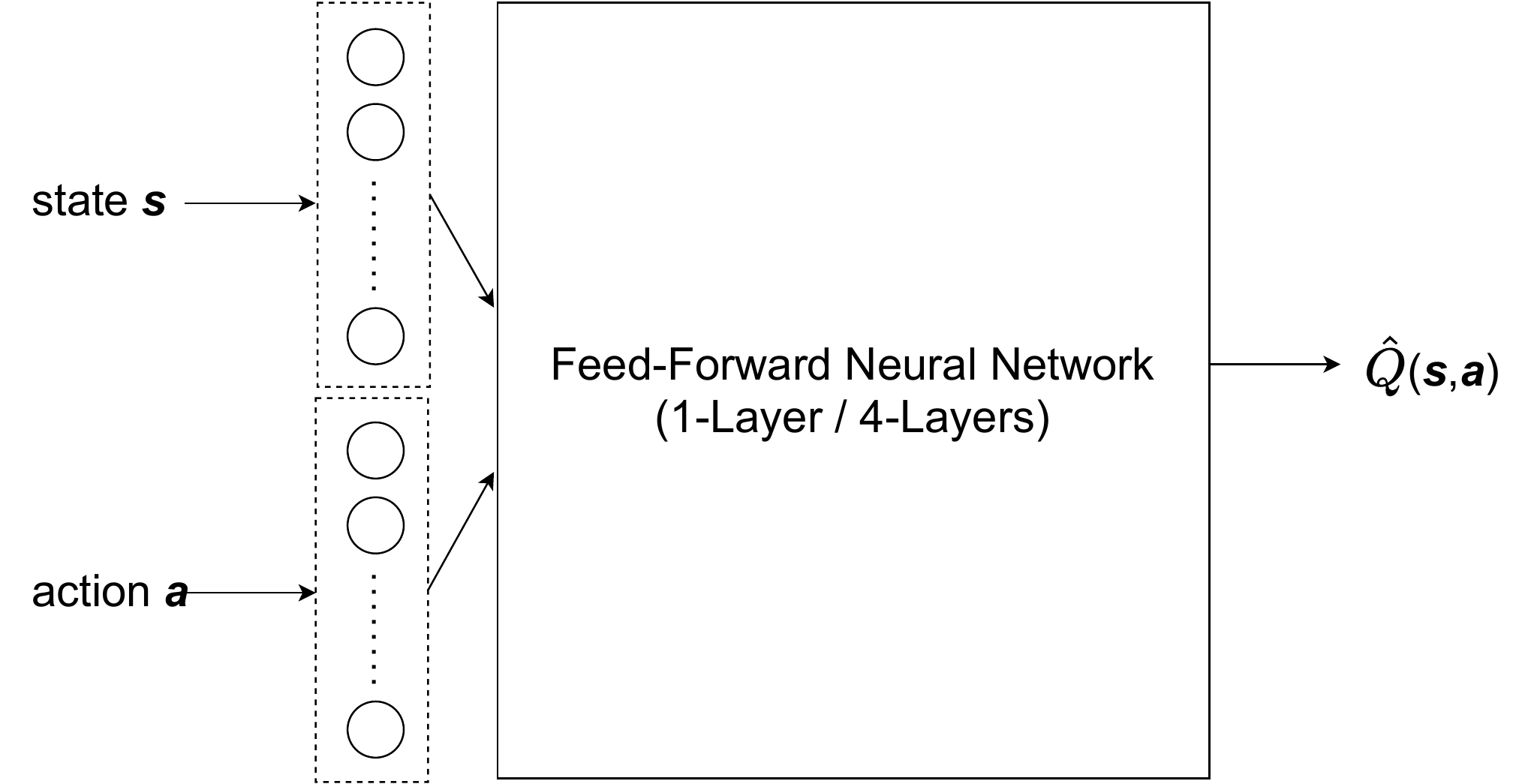}
	\vskip 0pt
	\caption{\small State-Action Approximator for $\ilbo$}
	\label{figS:Q_sa}
	\vskip 0pt
\end{figure}

\subsection{Hyperparameter Settings}

\begin{table}[th]
\begin{center}
\begin{tabular}{ | c | c | c | c | c | c |  }
\hline
\textbf{Neural Network Model} & \textbf{Activation} & \textbf{Normalization} & \textbf{Learning Rate} & \textbf{Sample Store Size} & \textbf{Minibatch Size} \\
\hline
State-Action Approximator & ReLU & Layer Normalization & 0.001 & 1M & 64 \\
DRP & ReLU & Layer Normalization & 0.0001 & 1K & 64 \\
\hline
\end{tabular}
\caption{Hyperparameter Settings for Neural Network Models of $\ilbo$}
\label{tableS:hyperparam}
\end{center}
\end{table}

The same two DRP architectures specified in $\tfmdp$ paper~\citeSupplement{ap_BuenoBMS19} were used in our empirical evaluation: one feed-forward neural network with 1-layer of hidden units [2048] and the other one with 4-layers of hidden units [256, 128, 64, 32]. In addition, $\ilbo$ learns another neural network as state-action approximator $\hat{Q}(s, a)$ to estimate the state value function $V^m(s)$. We retain the same architectures for this neural network, with an additional encoding layer before the feed-forward neural net. This encoding layer consists of 32 neurons for state vector and another 32 neurons for action vector, as illustrated in Figure~\figQ.

The hyperparameter settings for the two neural networks are tabulated in Table~\tabH.  
The same hyperparameter settings are used for all experiments across all three domains. Layer normalization~\citeSupplement{ap_ba2016layer} is used between hidden layers. The value of update rate $\tau$ is set to be 0.005 in our empirical experiments. 

\subsection{Hardware \& Software Specifications}

All experiments were conducted on an AMD EPYC 7371 16-Core CPU machine with 512 GB memory, running Ubnutu 18.04 LTS. The Python package of Tensorflow v1.15.5 was used to train the neural networks. The rddlgym~\citeSupplement{ap_rddlgym2020} version is v0.5.15 and $\tfmdp$~\citeSupplement{ap_tfmdp2021} version is v0.5.4.

\subsection{Additional Experiment Results}

We also verified ILBO's generalization ability in NAV-3 domain. Figure~\figI{}  
compares the same ILBO policy's solution quality against retrained $\tfmdp$ in new initial states. This again confirms that ILBO generalizes to new initial states without retraining.
\begin{figure*}[ht]
	\centering
	\includegraphics[width=1.0\textwidth]{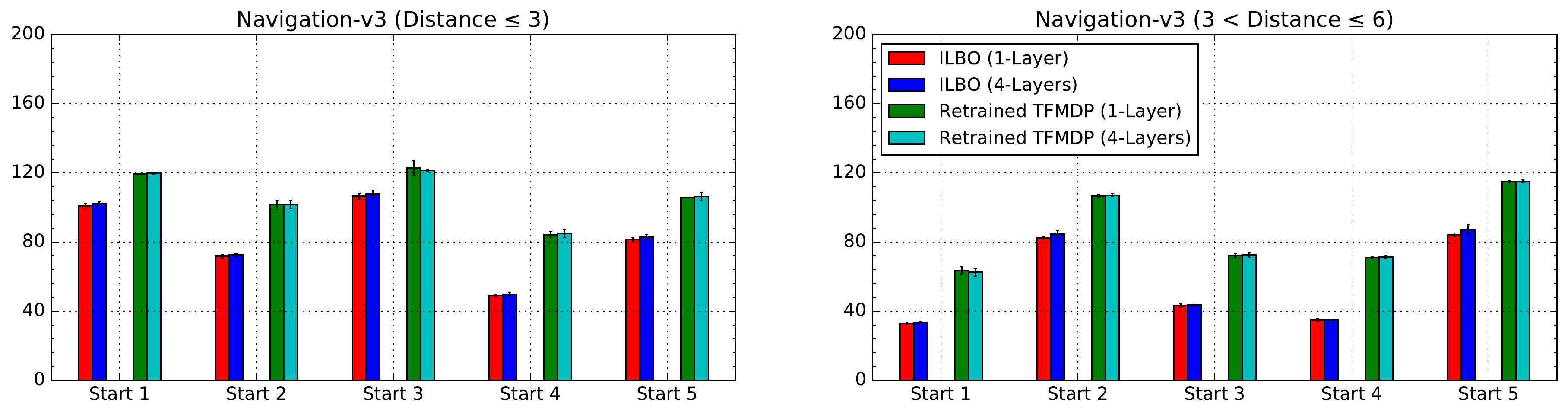}
	\caption{\small Total Cost Incurred in New Initial States (Lower is Better) for NAV3 Domain: Same $\ilbo$ Policy vs Retrained $\tfmdp$ Policy}
	\label{figS:NAV3_new_init_states}
\end{figure*}

\subsection{Code Appendix}

The source code used for conducting experiments on NAV3 domain is provided in the Code Appendix. The source code for the other two domains are almost identical, except some differences in their respective reward and transition functions. All source code will be made publicly available on GitHub after the review.

{
\small 
\bibliographystyleSupplement{aaai22}
\bibliographySupplement{aaai22MM-Appendix}}